\documentclass[sigconf]{acmart}

\usepackage{booktabs,multirow} 
\usepackage{hyperref}
\usepackage{amsmath}
\usepackage[ruled,vlined,linesnumbered]{algorithm2e}
\usepackage{soul}
\usepackage{bm}
\usepackage{enumitem}
\usepackage{tikz}
\newcommand{\tikzcircle}[2][red,fill=red]{\tikz[baseline=-0.5ex]\draw[#1,radius=#2] (0,0) circle ;}

\makeatletter 
\def\algbackskip{\hskip-\ALG@thistlm}
\makeatother 

\setcopyright{acmcopyright}

\acmDOI{10.1145/3377929.3398084}

\acmISBN{978-1-4503-7127-8/20/07}

\acmConference[GECCO '20]{the Genetic and Evolutionary Computation Conference 2020}{July 8--12, 2020}{Canc\'un, Mexico}
\acmYear{2020}
\copyrightyear{2020}

\acmPrice{15.00}

\begin{document}

\title[dMFEA-II for Discrete Optimization Problems]{dMFEA-II: An Adaptive Multifactorial Evolutionary Algorithm for Permutation-based Discrete Optimization Problems}


\author{Eneko Osaba}
\orcid{0000-0001-7863-9910}
\affiliation{%
	\institution{TECNALIA, Basque Research and Technology Alliance (BRTA)\\48160 Derio, Spain}
}
\email{eneko.osaba@tecnalia.com}

\author{Aritz D. Martinez}
\orcid{}
\affiliation{%
     \institution{TECNALIA, Basque Research and Technology Alliance (BRTA)\\48160 Derio, Spain}
}
\email{aritz.martinez@tecnalia.com}

\author{Akemi Galvez}
\orcid{}
\affiliation{%
	\institution{Universidad de Cantabria\\39005 Santander, Spain\\
	Toho University, Funabashi, Japan}
}
\email{galveza@unican.es}

\author{Andres Iglesias}
\orcid{}
\affiliation{%
	\institution{Universidad de Cantabria\\39005 Santander, Spain\\
		Toho University, Funabashi, Japan}
}
\email{iglesias@unican.es}

\author{Javier Del Ser}
\orcid{}
\affiliation{%
	\institution{University of the Basque Country (UPV/EHU), 48013 Bilbao, Spain}
}
\email{javier.delser@ehu.eus}

\renewcommand{\shortauthors}{E. Osaba et al.}



\begin{abstract}
The emerging research paradigm coined as multitasking optimization aims to solve multiple optimization tasks concurrently by means of a single search process. For this purpose, the exploitation of complementarities among the tasks to be solved is crucial, which is often achieved via the transfer of genetic material, thereby forging the Transfer Optimization field. In this context, Evolutionary Multitasking addresses this paradigm by resorting to concepts from Evolutionary Computation. Within this specific branch, approaches such as the Multifactorial Evolutionary Algorithm (MFEA) has lately gained a notable momentum when tackling multiple optimization tasks. This work contributes to this trend by proposing the first adaptation of the recently introduced Multifactorial Evolutionary Algorithm II (MFEA-II) to permutation-based discrete optimization environments. For modeling this adaptation, some concepts cannot be directly applied to discrete search spaces, such as parent-centric interactions. In this paper we entirely reformulate such concepts, making them suited to deal with permutation-based search spaces without loosing the inherent benefits of MFEA-II. The performance of the proposed solver has been assessed over 5 different multitasking setups, composed by 8 datasets of the well-known Traveling Salesman (TSP) and Capacitated Vehicle Routing Problems (CVRP). The obtained results and their comparison to those by the discrete version of the MFEA confirm the good performance of the developed dMFEA-II, and concur with the insights drawn in previous studies for continuous optimization.
\end{abstract}

\ccsdesc[500]{Theory of computation~Bio-inspired optimization}
\ccsdesc[300]{Theory of computation~Random search heuristics}
\ccsdesc[300]{Mathematics of computing~Evolutionary algorithms}
\ccsdesc[300]{Theory of computation~Theory of randomized search heuristics}

\keywords{Transfer Optimization, Evolutionary Multitasking, Multifactorial Optimization, Discrete Optimization, Traveling Salesman Problem}

\maketitle


\section{Introduction} \label{sec:intro}

The main motivation behind the recent Transfer Optimization paradigm is that real-world optimization problems hardly occur in isolation \cite{gupta2017insights}. Thus, the key idea on which this paradigm relies is the exploitation of what has been learned by optimizing one task when facing another problem or task. To tackle this paradigm, three different categories of Transfer Optimization can be distinguished: \textit{sequential transfer}, \textit{multitasking} and \textit{multiform optimization} \cite{gupta2017insights,feng2015memes}. Among these three classes, \textit{multitasking} is arguably the one that has attracted most attention by the current community \cite{gupta2016genetic,wen2017parting}, which is devoted to simultaneously solving different optimization problems or tasks by dynamically analyzing existing synergies and complementarities among them. 

Given the context above, this manuscript is focused on Evolutionary Multitasking (EM, \cite{ong2016towards}), a branch of Transfer Optimization that relies on concepts from Evolutionary Computation for the simultaneous solving of different problems \cite{back1997handbook,del2019bio}. In the last few years, several EM proposals have been reported in the literature to deal with several discrete, continuous, single-objective and multi-objective optimization problems at the same time \cite{wang2019evolutionary,gong2019evolutionary,yu2019multifactorial,gupta2016multiobjective}. From the algorithmic point of view, in most of the aforementioned studies EM has been materialized by means of the so-called Multifactorial Optimization (MFO) strategy, which hinges on the definition of a unique factor for each individual to influence the search of population-based solvers. Most notably, the combination of MFO and EM has given rise to the Multifactorial Evolutionary Algorithm (MFEA, \cite{gupta2015multifactorial}), arguably at the forefront of the algorithms contributed so far in the area.

Despite the relative youth of the field, there is a clear consensus in the community about the paramount importance of the correlation among tasks to be simultaneously addressed. Exploiting this correlation is crucial in order to positively leverage the transfer of knowledge over the search \cite{zhou2018study}. Several influential contributions can be found in the literature delving into this issue, proposing alternatives to analyze and quantify the similarity between optimization tasks \cite{gupta2016landscape}. Indeed, in practical setups it is not possible to ensure that all tasks are related to each other. In such cases, overlooking this lacks of synergy, and sharing genetic material among unrelated tasks or problems could lead to performance downturns, a circumstance known as \textit{negative transfer} \cite{bali2019multifactorial}. This negative transfer has been reported by some recent studies as the central pitfall of multitasking, becoming a priority in the formulation of new schemes \cite{zhang2019multi,chen2019adaptive}. Among them, the brand new Multifactorial Evolutionary Algorithm II (MFEA-II, \cite{bali2019multifactorial}) is an adaptive extension of the aforementioned MFEA, incorporating the capability to dynamically learn how much knowledge should be transferred across tasks. 

As evinced by the literature so far, MFEA-II has so far been tested over continuous optimization problems, using experimental environments composed by up to $6$ tasks. The lack of applications with alternative problem flavors, and wider experimental setups, comprise the main source of motivation of this research work. Specifically, we elaborate on adapting MFEA-II to permutation-based combinatorial problems, giving rise to the discrete MFEA-II (dMFEA-II). Despite the simple formulation of our research hypothesis, the adaptation beneath dMFEA-II is not straightforward, as the naive version of MFEA-II is comprised by concepts and operators that cannot be directly applied to permutation-based discrete search spaces. An example supporting this statement is the parent-based strategy followed for the inter-task interactions \cite{deb2002real}, or the transfer parameter matrix, crucial for the search procedure of MFEA-II. We assess the performance of our proposed dMFEA-II by considering 8 instances of the well-known Traveling Salesman (TSP, \cite{lawler1985traveling}) and Capacitated Vehicle Routing (CVRP, \cite{ralphs2003capacitated}) problems, which are combined to yield 5 multitasking environments with heterogeneous search spaces and varying degrees of phenotypical relationship. Results obtained by dMFEA-II are compared to those of the discrete version of the MFEA, aimed at the confirmation of the same findings drawn from \cite{bali2019multifactorial} for continuous optimization environments.

The remainder of the article is organized as follows. Section \ref{sec:back} sets the background and related work. Next, Section \ref{sec:MFEAII} exposes in detail the main features of the proposed dMFEA-II. The experimentation setup and discussion of the results are given in Section \ref{sec:exp}. Finally, Section \ref{sec:conc} concludes the paper.

\section{Background and Related Work} \label{sec:back}

There is global consensus in the community that until 2017, the concept of EM was only formulated within the framework of MFO \cite{da2017evolutionary}. In the last years, several approaches have embraced this concept \cite{xiao2019multifactorial,yokoya2019multifactorial,osaba2020multifactorial,li2020multifactorial}, with MFEA at the spearhead \cite{gupta2015multifactorial}. Additional alternatives to MFO have been also proposed in terms of new algorithmic schemes, such as the multitasking multi-swarm optimization in \cite{song2019multitasking}, or the coevolutionary multitasking schemes in \cite{cheng2017coevolutionary,osaba2020coeba}. 

Deeper into mathematical details, MFO can be formulated by considering an environment comprising $K$ tasks or problems to be simultaneously solved. This environment is therefore made up by as many search spaces as tasks to be faced. Therefore the objective function for the $k$-th task $T_k$ is denoted as $f_k : \Omega_k \rightarrow \mathbb{R}$, where $\Omega_k$ is the search space of task $T_k$. Assuming that all tasks should be minimized, the main objective is to find a group of solutions $\{\mathbf{x}_1,\dots,\mathbf{x}_K\}$ such that $\mathbf{x}_k = \arg \min_{\mathbf{x}\in\Omega_k} f_k(\mathbf{x})$. In general, a MFO algorithm operates on a population $\mathcal{P}$ of candidate solutions (individuals), where each $\mathbf{x}_p\in\mathcal{P}$ should belong to a unified search space $\Omega_U$. Each search space $\Omega_k$ is mapped to $\Omega_U$ through the use of an encoding/decoding function $\xi_k: \Omega_k\mapsto \Omega_U$. Consequently, every individual $\mathbf{x}_p\in\mathcal{P}$ should be encoded as $\mathbf{x}_{p,k}=\xi_k^{-1}(\mathbf{x}_p)$ to represent a task-specific solution $\mathbf{x}_{p,k}$ for each of the $K$ tasks. Departing from these definitions, in every MFO solver four different features are associated with each individual $\mathbf{x}_p$ of the population $\mathcal{P}$: \textit{Factorial Cost}, \textit{Factorial Rank}, \textit{Scalar Fitness} and \textit{Skill Factor}. These features permit to sort, select and/or discard individuals along the search, as they dictate the contribution of every individual to the population considering that $K$ tasks are optimized \cite{bali2019multifactorial}:
\begin{itemize}[leftmargin=*]
\item \textit{Factorial Cost} $\Psi_p^k$ of an individual $\mathbf{x}_p\in\mathcal{P}$ is given by its fitness value for task $T_k$, so that each solution in the population retains a list \smash{$\{\Psi_p^1,\Psi_p^2,\dots,\Psi_p^K\}\in\mathbb{R}^K$} of factorial costs.

\item \textit{Factorial Rank} $r_p^k$ of an individual $\mathbf{x}_p$ in a given task $T_k$ is its relative rank within the population in ascending order of \smash{$\Psi_p^k$}. Similarly to the factorial cost, each individual can be characterized by a factorial rank list $\{r_p^1,r_p^2,\dots,r_p^K\}\in\mathbb{N}^K$.

\item \textit{Scalar Fitness} $\varphi_p$ of $\mathbf{x}_p$ is given by its best factorial rank over all tasks as \smash{$\varphi_p = 1/ \min_{k \in \{1...K\}}r_p^k$}. The scalar fitness permits to compare different individuals in MFEA.

\item The \emph{Skill Factor} $\tau_p$ is the task in which $\mathbf{x}_p$ performs best, namely, \smash{$\tau_p = \arg \min_{k\in\{1,\ldots,K\}} r_p^k$}. As we will show later, the skill factor plays a crucial role in MFEA by establishing which population members are selected for crossover.
\end{itemize}

When operating on the population of individuals via evolutionary methods, EM emerges as an effective paradigm for tackling multiple problems simultaneously. This efficiency is due to i) the parallelism granted by having a population of individuals, which eases the concurrent application of evolutionary operators and the dynamic estimation of latent synergies between tasks \cite{ong2016evolutionary}; and ii) the exchange of genetic material among individuals through crossover methods, allowing all tasks to interact with each other. Among them, the specific MFEA approach is based on bio-cultural schemes of multifactorial inheritance. We depict in Algorithm \ref{alg:classicMFEA} the pseudo-code of the basic MFEA, which has four key characteristics:
\begin{itemize}[leftmargin=*]
	\item \textit{Unified search space}: one of the main design challenges when modeling a MFEA is the definition a unified space $\Omega_U$, which should be able to represent all feasible solutions of the $K$ tasks. 
	
	\item \textit{Assortative mating}, which is based on the principle that individuals are more inclined to interact with others belonging to the same cultural background. For this reason, genetic operators used in MFEA are committed to follow this principle, promoting interactions among solutions with the same skill factor. We again recommend \cite{gupta2015multifactorial} for more details on this procedure.
	
	\item \textit{Selective evaluation}: every newly created individual is measured only on one task. This procedure guarantees the computational feasibility of the method. Specifically, each new solution is evaluated in the task corresponding to the skill factor of its parent. When mating two parents, the skill factor of the offspring is selected randomly among those of the parents. 
	
	\item \textit{Scalar fitness based selection}, which can be conceived as an elitist replacement strategy that uses the scalar fitness (namely, the best relative rank of the individual over all tasks) as the control parameter. In other words, the best $P$ solutions (considering both newly generated individuals and the current population) in terms of scalar fitness survive for the next generation.
\end{itemize}
\begin{algorithm}[h!]
	\SetAlgoLined
	\DontPrintSemicolon
	Randomly draw a population of $|\mathcal{P}|=P$ individuals $\{\mathbf{x}_p\}_{p=1}^P$, with $\mathbf{x}_p\in\Omega_U$\;
	Evaluate each generated individual for the $K$ problems\;
	Calculate the skill factor $\tau_p$ of each $\mathbf{x}_p$\;
	\While{termination criterion not reached}{
		Set $\mathcal{Q}=\emptyset$\;
		\While{individuals still to select}{
			Randomly sample w/out replacement $\mathbf{x}_{p'},\mathbf{x}_{p''}\in\mathcal{P}$\;
			\uIf{$\tau_{p'}=\tau_{p''}$}{
				$[\mathbf{x}_{A},\mathbf{x}_{B}] = \mbox{IntrataskCX}(\mathbf{x}_{p'},\mathbf{x}_{p''})$\;
				Set $\tau_A$ and $\tau_B$ equal to $\tau_{p'}$\;
			}\uElseIf{$rand\leq \mbox{RMP}$}{
				$[\mathbf{x}_{A},\mathbf{x}_{B}] = \mbox{IntertaskCX}(\mathbf{x}_{p'},\mathbf{x}_{p''})$\;
				Set $\tau_A=rand(\tau_{p'},\tau_{p''})$ and $\tau_B=rand(\tau_{p'},\tau_{p''})$\;
			}\uElse{
				Compute $\mathbf{x}_A = \mbox{mutation}(\mathbf{x}_{p'})$, and set \smash{$\tau_A=\tau_{p'}$}\;
				Compute $\mathbf{x}_B = \mbox{mutation}(\mathbf{x}_{p''})$, and set \smash{$\tau_B=\tau_{p''}$}\;
			}
			Evaluate $\mathbf{x}_A$ for task $\tau_A$, and $\mathbf{x}_B$ for task $\tau_B$\;
			$\mathcal{Q}$ = $\mathcal{Q}\cup\{\mathbf{x}_A,\mathbf{x}_B\}$\;
		}
		Select the best $P$ individuals in $\mathcal{P}\cup\mathcal{Q}$ as per their $\varphi_p$ \;
	}
	Return the best individual in $\mathcal{P}$ for each task $T_k$\;
	\caption{Pseudocode of MFEA}
	\label{alg:classicMFEA}
\end{algorithm}

In recent years, a manifold of contributions have been inspired by the algorithmic principles of MFEA. In \cite{zhou2016evolutionary}, for example, a discrete adaptation of the canonical MFEA is proposed and applied to one of the problems addressed on this present paper: the CVRP. The research work in \cite{yuan2016evolutionary} also goes in the same direction by introducing the discrete unified encoding, which has thereafter served as a reference when dealing with different discrete problems via MFEA. Furthermore, authors in \cite{gong2019evolutionary} implemented an improved variant of the MFEA, endowing the meta-heuristic with a dynamic resource allocating strategy. A similar solver is presented in \cite{yao2020multiobjective} for dealing with multiobjective optimization tasks. Other applications of MFEA can be found in \cite{wang2019evolutionary} for the composition of semantic web services, and in \cite{martinez2020simultaneously} for evolving deep reinforcement learning models. 

Despite this success, MFEA, EM and the wider field of Transfer Optimization are also in the focus of few critical researchers, who question the operation of the methods implemented so far. Mainly, these skeptical voices refer to the difficulty of avoiding negative transfers and reacting to their existence \cite{wang2019rigorous}. In fact, it is well accepted that the performance of Transfer Optimization algorithms is directly related to the synergies between the problems involved \cite{ong2016towards,zhou2018study,da2017evolutionary}. For this reason, the community is striving to propose new methods to cope with this situation, favoring positive transfers, and making optimization algorithms adaptive to avoid negative influences among tasks \cite{cai2019multitasking,liang2020two}. This is in fact the main purpose of the recently proposed adaptive variant of MFEA, coined as MFEA-II \cite{bali2019multifactorial}. MFEA-II introduces new algorithmic ingredients that make its search resilient against negative information transfer. The next section describe these ingredients and their adaptation for efficiently solving permutation-based problems.

\section{Proposed \lowercase{d}MFEA-II Approach}\label{sec:MFEAII}

The main novelty introduced by MFEA-II with respect to its predecessor is the introduction of a transfer parameter matrix, which dictates the way in which the inter-tasks relationships are conducted, and whose entries are evolved based on the information generated during the course of the multitasking search. In accordance with the claims in \cite{bali2019multifactorial}, the initial phases of MFEA-II are the same as those in MFEA. With this, the main differential factor is the incorporation of the online RMP learning module, and its foundry within the optimization process. This learning module is in charge of building and managing the dynamic RMP matrix, which dictates the extent of genetic transfer across individuals with different \textit{skill task} (see line 11 in Algorithm \ref{alg:classicMFEA}). Another feature of MFEA-II is the inter-task crossover procedure, activated when individuals with different skill tasks should interact, and fully conducted using parent-centric operators \cite{deb2002real}. In other words, lines 11-16 in Algorithm \ref{alg:classicMFEA} are replaced by those in Algorithm \ref{alg:MFEAIImodifiedsteps}. The main contribution of the present paper is specifically the adaptation of these steps in order to deal with discrete optimization environments. 
\begin{algorithm}[h!]
	\SetAlgoLined
	\DontPrintSemicolon
	\uIf{$\tau_{p'}\neq\tau_{p''}$}{
		\uIf{$rand\leq \mbox{RMP}_{\tau_{p'},\tau_{p''}}$}{
			$[\mathbf{x}_A,\mathbf{x}_B] = \mbox{IntertaskParentCentricCX}(\mathbf{x}_{p'},\mathbf{x}_{p''})$\;
			Update $\mathbf{x}_A = \mbox{mutation}(\mathbf{x}_A)$, and $\mathbf{x}_B = \mbox{mutation}(\mathbf{x}_B)$\;
			Set $\tau_A=rand(\tau_{p'},\tau_{p''})$ and $\tau_B=rand(\tau_{p'},\tau_{p''})$\;
		}\uElse{
			Randomly select $\mathbf{x}_{p1}\in\mathcal{P}$ with $\tau_{p1}=\tau_{p'}$ and $p1\neq p'$\;
			$\mathbf{x}_A = \mbox{IntrataskParentCentricCX}(\mathbf{x}_{p'},\mathbf{x}_{p1})$\;
			Update $\mathbf{x}_A = \mbox{mutation}(\mathbf{x}_A)$, and set $\tau_A=\tau_{p'}$\;
			Randomly select $\mathbf{x}_{p2}\in\mathcal{P}$ with $\tau_{p2}=\tau_{p''}$ and $p2\neq p''$\;
			$\mathbf{x}_B = \mbox{IntertaskParentCentricCX}(\mathbf{x}_{p''},\mathbf{x}_{p2})$\;
			Update $\mathbf{x}_B = \mbox{mutation}(\mathbf{x}_B)$, and set $\tau_B=\tau_{p''}$\;
		}
	}
	\caption{Inter-task crossover procedure of MFEA-II}
	\label{alg:MFEAIImodifiedsteps}
\end{algorithm}

Before proceeding further, we now pause at the main rationale for the need of this adaptation. As mentioned previously, the search process of MFEA-II hinges on parent-centric operators, such as the Simulated Binary Crossover \cite{deb1995simulated}, the Polynomial Mutation \cite{deb1996combined} or the Gaussian Mutation \cite{hinterding1995gaussian} with small variance. These operators are known to produce individuals close to their parents in the unified search space $\Omega_U$. All these operators were originally conceived for continuous optimization problems \cite{deb2002computationally,garcia2008global}, and have no clear correspondence for discrete problems as the permutation-based ones considered in this work. 

The second issue when adapting the canonical MFEA-II to combinatorial optimization problems is the mutation mechanism in use. While in MFEA local perturbations are only conducted when $\tau_{p'}\neq\tau_{p''}$ and $rand> \mbox{RMP}$ (lines 15 and 16 of Algorithm \ref{alg:classicMFEA}), in MFEA-II \textit{each generated offspring also undergoes small parent-centric mutation} \cite{bali2019multifactorial}. This new procedure requires a reformulation when solving permutation-based problems, in which operators such as 2-opt, 3-opt, swapping or insertion \cite{johnson1997traveling,larranaga1999genetic} involve a small change in the individual. This small perturbation, along with the previous crossover, could lead to drastically modified $\mathbf{x}_A$ and $\mathbf{x}_B$ individuals. This potentially intensified change of the produced offspring clashes with the main search behavior of MFEA-II. This same trend also holds for the dynamic RMP matrix, which has also been adapted to the typology of problems addressed in this paper.
\begin{algorithm}[h!]
	\SetAlgoLined
	\DontPrintSemicolon
	\uIf{$\tau_{p'}\neq\tau_{p''}$}{
		\uIf{\smash{$rand1\leq \mbox{RMP}_{\tau_{p'},\tau_{p''}}$}}{
			$[\mathbf{x}_A,\mathbf{x}_B] = \mbox{IntertaskParentCentricCX}(\mathbf{x}_{p'},\mathbf{x}_{p''})$\;
			\uIf{$rand2<P_m$}{ 
				$\mathbf{x}_A = \mbox{mutation}(\mathbf{x}_A)$; $\mathbf{x}_B = \mbox{mutation}(\mathbf{x}_B)$\;
			}
			Set $\tau_A=rand(\tau_{p'},\tau_{p''})$ and $\tau_B=rand(\tau_{p'},\tau_{p''})$\;
			Update \smash{$\mbox{RMP}_{\tau_{p'},\tau_{p''}}$} using $\Delta_{inc}$ or $\Delta_{dec}$\;
		}\uElse{
			Randomly select $\mathbf{x}_{p1}\in\mathcal{P}$ with $\tau_{p1}=\tau_{p'}$ and $p1\neq p'$\;
			$\mathbf{x}_A = \mbox{IntrataskParentCentricCX}(\mathbf{x}_{p'},\mathbf{x}_{p1})$\;
			\uIf{$rand2\leq P_m$}{
				$\mathbf{x}_A = \mbox{mutation}(\mathbf{x}_A)$\;
			}
			Set $\tau_A=\tau_{p'}$, and update \smash{$\mbox{RMP}_{\tau_{p'},\tau_{p'}}$}\;
			Randomly select $\mathbf{x}_{p2}\in\mathcal{P}$ with $\tau_{p2}=\tau_{p''}$ and $p2\neq p''$\;
			$\mathbf{x}_B = \mbox{IntrataskParentCentricCX}(\mathbf{x}_{p''},\mathbf{x}_{p2})$\;
			\uIf{$rand2\leq P_m$}{
				$\mathbf{x}_B = \mbox{mutation}(\mathbf{x}_B)$\;
			}
			Set $\tau_B=\tau_{p''}$, and update \smash{$\mbox{RMP}_{\tau_{p''},\tau_{p''}}$}\;
		}
	}
	\caption{Crossover strategy of dMFEA-II}
	\label{alg:dMFEAII}
\end{algorithm}

In light of the above, Algorithm \ref{alg:dMFEAII} summarizes the scheme proposed in our dMFEA-II for the inter-task crossover procedure, which replaces lines 11-16 in Algorithm \ref{alg:classicMFEA} and the whole pseudocode depicted in Algorithm \ref{alg:MFEAIImodifiedsteps}. To begin with, a permutation encoding is employed as unified representation $\Omega_U$ for $\mathcal{P}$, as also done in other studies \cite{yuan2016evolutionary,zhou2016evolutionary}. Having said this, if $K$ problems are to be addressed, and representing the dimensionality of each instance $T_k$ as $D_k\in\mathbb{N}$, a solution $\mathbf{x}_p$ is represented as a permutation of the integer set $\{1,2,\ldots, D_{max}\}$, where $D_{max}=\max_{k\in\{1,\ldots,K\}} D_k$ (maximum dimension among the $K$ tasks). Hence, if an individual $\mathbf{x}_p^\prime$ is going to be measured on a task $T_k$ whose $D_k<D_{max}$, only integers lower than $D_k$ are considered for producing the solution $\mathbf{x}_k$ of $T_k$. We now describe the main modifications conducted over the basic MFEA-II in order to properly face permutation-based problems:
\begin{enumerate}[wide,leftmargin=0pt]
	\itemsep0.2em 
	\item[1)] First of all, dMFEA-II implements \emph{a simple strategy for dynamically adapting the RMP matrix} to the search performance. Following the philosophy of the online RMP learning module for continuous scenarios described in \cite{bali2019multifactorial}, we have designed a reliable alternative strategy, simple but effective, to dictate the intensity and frequency of the interactions of tasks of different kind. First, as for the continuous MFEA-II, in our dMFEA-II RMP is not a single parameter but a symmetric $K \times K$ matrix, with $K$ denoting the number of optimization tasks. The entries of this matrix are real-valued in the range $[0.0,1.0]$, so that $\mbox{RMP}_{k,k'}$ indicates the probability of conducting an inter-task crossover between tasks $k$ and $k'$. All $\mbox{RMP}_{k,k'}$ are initially set to a relatively high value (e.g. $0.95$) in order to facilitate all task interactions in the initial stages of the search. Furthermore, two additional control parameters are defined: $\Delta_{dec}$ and $\Delta_{inc}$. These parameters are set to a real value withing the interval $[0.0,1.0]$, and determine the evolution of each $\mbox{RMP}_{k,k'}$ entry in the following manner: each time a new individual is created (e.g. $\mathbf{x}_A$ as per lines 3 to 5 of Algorithm \ref{alg:dMFEAII}), its factorial cost is calculated and compared to the parent $\mathbf{x}_{p'}$ from which its \textit{skill task} $\tau_A$ has been inherited. In case $\mathbf{x}_A$ obtains a better performance in the \textit{skill task} of its parent, we can ensure that the genetic transfer between tasks $\tau_{p'}$ and $\tau_{p''}$ has been positive. Thus, $\mbox{RMP}_{\tau_{p'},\tau_{p''}}$ is incremented using $\Delta_{inc}$ parameter control as $\mbox{RMP}_{\tau_{p'},\tau_{p''}}=\min\{1.0,\mbox{RMP}_{\tau_{p'},\tau_{p''}}/\Delta_{inc}\}$. Otherwise, we can categorize the transfer as negative, decrementing the value of $\mbox{RMP}_{\tau_{p'},\tau_{p''}}$ as $\mbox{RMP}_{\tau_{p'},\tau_{p''}}=\max\{0.1,\mbox{RMP}_{\tau_{p'},\tau_{p''}}\cdot\Delta_{dec}\}$. A lower bound of $\mbox{RMP}_{\tau_{p'},\tau_{p''}}$ is set to maintain a minimum knowledge exchange between any two tasks. Lastly, the intra-task crossover conducted in dMFEA-II if $rand > \mbox{RMP}_{\tau_{p'},\tau_{p''}}$ are also parent-centric, so that the evaluation and comparison of the produced individuals update $\mbox{RMP}_{\tau_{p'},\tau_{p''}}$ and $\mbox{RMP}_{\tau_{p'},\tau_{p''}}$ by following the previous rules.
	
	\item[2)] In order to counteract the aforementioned intensification of changes imprinted to the offspring, dMFEA-II introduces a \emph{mutation parameter} $P_m\in[0,1]$ to control whether a new individual $\mathbf{x}_A$ or $\mathbf{x}_B$ should undergo mutation. 
	
	\item[3)] The \textit{dynamic discrete parent-centric operator for both inter-task and intra-task crossover} designed for the dMFEA-II is based on the fulfillment of two different considerations. The first one is its parent-centric nature. In other words, created individuals should not be far away with respect to their parents (a \emph{small leap} in the search space). This first consideration can be realized by just fixing one of the parents as dominant, and limiting the amount of genetic material transferred from the other parent. The second factor is the dynamic nature of the operator. By virtue of this feature, the crossover function adapts its operation to the synergies arisen between optimization tasks over the search. This entails that if the complementarity shown among tasks $k$ and $k'$ is high, the amount of genetic material transferred between these tasks should also be high, and vice versa. In this way, since the RMP matrix should dynamically reflect the effectiveness of knowledge sharing between tasks, we use the values in this matrix for materializing the dynamic parent-centric characteristic of the crossover in dMFEA-II. 
	
	Without loss of generality we consider the Order Crossover (OX, \cite{davis1985job}) to exemplify how we translated this concept to the specific case study presented in this paper. The main principles of OX is to randomly choose two different cutting points in the problem solution, in order to define the segment of the individual (\emph{cutting window}) that decides the amount of genetic material transferred from one parent to another. The first change done to adapt OX to this parent-centric feature is to limit the size of the cutting window to a fraction $W\in[0,1]$ of the total dimension. This maximum size, along with the value of $\mbox{RMP}_{k,k'}$, would set the amount of genetic material transferred from task $T_k$ to $T_{k'}$ as $W\cdot \mbox{RMP}_{k,k'}\cdot D_k$, where $D_k$ is the dimensionality of task $T_k$. Namely, for tasks with a fully positive synergy in terms of knowledge transfer ($\mbox{RMP}_{k,k'}=1.0$), the size of the shared material would be equal to $W\cdot D_k$. Finally, if the amount of elements transferred is so low that it is not possible to ensure a variability between the parent and the generated child, a $2-opt$ mutation is conducted to ensure that offspring and parents differ. We have coined this modified crossover operator as Dynamic OX (dOX), which will be later used in the experimental part of the study. Depending on the problems under consideration, other crossover functions could be also considered and reformulated to incorporate the dynamic and parent-centric nature of dMFEA-II.
\end{enumerate}

\section{Experimentation and Results}\label{sec:exp}

In order to shed light over the performance of the proposed dMFEA-II approach, an experimental benchmark has been designed considering both TSP and CVRP instances to be simultaneously optimized. Readers interested on these classical problems are referred to recent surveys such as \cite{caceres2015rich,osaba2020vehicle}. In particular, we assess the efficiency of dMFEA-II and its MFEA counterpart over 8 TSP and CVRP instances, which are combined to yield 5 different tests scenarios. All TSP instances have been obtained from the TSPLIB repository \cite{TSPLib}: \texttt{berlin52}, \texttt{eil51}, \texttt{st70} and \texttt{eil76}. Sizes of these instances are 52, 51, 70 and 76, respectively. On the other hand, the CVRP instances are part of the Augerat Benchmark \cite{augerat1995computational}: \texttt{P-n50-k7}, \texttt{P-n50-k8}, \texttt{P-n55-k7} and \texttt{P-n55-k8}. The dimensions of these cases are 50 in the first two datasets and 55 in the remaining two. We have opted for related instances, as e.g. all the CVRP or \texttt{eil51}-\texttt{eil76}; and non-related instances, such as \texttt{berlin52} and \texttt{st70} or any TSP instance compared to a CVRP one. In this way, we ensure that when facing the experimentation environments, dMFEA-II deals with both positive and negative sharing of knowledge. 
\begin{table}[h!]
	\centering
	\resizebox{0.9\columnwidth}{!}{
		\begin{tabular}{lcclc}
			\toprule 
			\multicolumn{2}{c}{dMFEA-II} & & \multicolumn{2}{c}{MFEA} \\
			\cmidrule{1-2} \cmidrule{4-5}
			Parameter & Value & & Parameter & Value\\
			\cmidrule{1-2} \cmidrule{4-5}
			Population size & 200 & & Population size & 200 \\ 
			Intra-task CX & OX \cite{davis1985job} & & CX & OX \\ 
			Mutation & 2-opt \cite{lin1965computer} & & Mutation & 2-opt \\
			Initial values of $RMP_{k,k'}$ & 0.95 & & RMP & 0.9  \\
			Parent-centric CX & Dynamic OX (dOX) & & & \\
			$P_m$ & 0.2 & &  & \\
			$\Delta_{inc}$ / $\Delta_{dec}$ & 0.99 / 0.99 & & & \\ \bottomrule
		\end{tabular}
	}
	\vspace{2mm}
	\caption{Parameter values for dMFEA-II and MFEA.}
	\vspace{-5mm}
	\label{tab:Parametrization}
\end{table}

Thus, 5 different multitasking environments have been constructed for the tests. Each of these scenarios implies that both dMFEA-II and MFEA should solve all the datasets assigned to the environment simultaneously. The main criterion for generating these particular environments is twofold: i) to exploit the possible genetic synergies of the instances and analyze the reaction of the dMFEA-II to negative interactions, and ii) to reach significant findings over a diverse group of multitasking environments, involving each TSP and CVRP instance in exactly the same number of environments. Four of these environments are composed by 4 different instances, while the last one, namely \emph{TE\_8}, contemplates all the 8 problem instances. \emph{TE\_4\_1} is comprised by the four TSP datasets, while the four CVRP datasets are included in \emph{TE\_4\_2}. The rest of multitasking setups comprise both TSP and CVRP instances. First, \emph{TE\_4\_3} is composed by \texttt{eil51}, \texttt{berlin52}, \texttt{P-n50-k7} and \texttt{P-n50-k8}. Lastly, \emph{TE\_4\_4} consists of \texttt{st70}, \texttt{eil76}, \texttt{P-n55-k7} and \texttt{P-n55-k8}.

For the sake of reproducibility of this research work, parameters employed for the developed methods are summarized in Table \ref{tab:Parametrization}. Some of these parameters, such as the population size $P$, $W$ (cutting window size), $\Delta_{inc}$ or $\Delta_{dec}$, have been tuned after an exhaustive search process not shown for lack of space. Other parameter values have been set as per other related works \cite{yuan2016evolutionary,gupta2015multifactorial,bali2019multifactorial}. All individuals in the population $\mathcal{P}$ have been initialized uniformly at random. As termination criterion, each solver finished its execution after $I=6\cdot 10^5$ objective function evaluations. To deal with CVRP problems, a set of zeros are dynamically inserted in the solution as control integers, with the aim of meeting the capacity constraints. Each multitasking configuration has been run 20 times to account for the statistical significance of performance gaps found along the tests. Lastly, all the experimentation has been conducted on an Intel Xeon E5-2650 v3 computer, with 2.30 GHz and a 32 GB RAM.

\subsection{Results and Discussion}\label{sec:exp_res}

Table \ref{tab:results} summarizes the outcomes attained by both dMFEA-II and MFEA for all the five test environments described above. Specifically, the table shows the average and standard deviation (computed over 20 independent runs) of the fitness obtained for each instance and multitasking configuration. Moreover, we provide the known optima for each TSP and CVRP instance. However, it is important to set clear, at this point, that the objective of the designed experimental benchmark is not to reach the optimal solution of the instances under consideration, but rather to use them as a reference of the performance of the designed multitasking approach.

\begin{table}[h!]
	\centering
	\renewcommand{\arraystretch}{1.1}
	\resizebox{\columnwidth}{!}{
		\begin{tabular}{cccccccccc}
			\toprule
			& \multirow{2}{*}{Method} & \multicolumn{4}{c}{TSP instances} & \multicolumn{4}{c}{CVRP instances} \\
			\cmidrule(lr){3-6} \cmidrule(lr){7-10} 
			& & \texttt{berlin52} & \texttt{eil51} & \texttt{st70} & \texttt{eil51} & \texttt{P-n50-k7} & \texttt{P-n50-k8} & \texttt{P-n55-k7} & \texttt{P-n55-k8} \\ \midrule
			            \multirow{5}{*}{\rotatebox{90}{\emph{TE\_4\_1}}} & 
			            \multirow{2}{*}{dMFEA-II} 
			            & \textbf{8078.8} & 450.3 & \textbf{721.2} & \textbf{585.1} & -- & -- & -- & --\\ 
			            & & 264.56 & 8.31 & 13.41 & 14.08 & -- & -- & -- & --\\ 
			            \cmidrule{2-10}
			            & \multirow{2}{*}{MFEA} 
			             & 8130.3 & \textbf{447.5} & 747.7 & 597.0 & -- & -- & -- & --\\ 
			            & & 275.51 & 5.14 & 21.17 & 8.23 & -- & -- & -- & --\\  
			            \cmidrule{2-10}
			            & Wilcoxon test & \tikzcircle[fill=white]{4pt} & \tikzcircle[fill=gray]{4pt} & \tikzcircle[fill=gray]{4pt} & \tikzcircle[fill=white]{4pt} & -- & -- & --& -- \\	
			            \midrule
			            
			            \multirow{5}{*}{\rotatebox{90}{\emph{TE\_4\_2}}} & 
			            \multirow{2}{*}{dMFEA-II} 
			            & -- & -- & -- & -- & \textbf{607.6} & \textbf{696.5} & \textbf{645.3} & 644.5\\ 
			            & & -- & -- & -- & -- & 20.59 & 19.26 & 17.48 & 27.10\\ 
			            \cmidrule{2-10}
			            & \multirow{2}{*}{MFEA} 
			            & -- & -- & -- & -- & 616.2 & 698.3 & 647.8 & \textbf{643.3} \\
			            & & -- & -- & -- & -- & 16.60 & 14.31 & 24.72 & 24.57 \\ 
			            \cmidrule{2-10}
			            & Wilcoxon test & -- & -- & -- & -- & \tikzcircle[fill=white]{4pt} & \tikzcircle[fill=gray]{4pt} & \tikzcircle[fill=gray]{4pt} & \tikzcircle[fill=gray]{4pt} \\	
			            \midrule
			            
			            \multirow{5}{*}{\rotatebox{90}{\emph{TE\_4\_3}}} & 
			            \multirow{2}{*}{dMFEA-II} 
			            & \textbf{8151.8} & \textbf{447.8} & -- & -- & \textbf{628.8} & \textbf{704.2} & -- & --\\ 
			            & & 229.95 &  9.28 & -- & -- & 15.27 & 30.06 & -- & --\\ 
			            \cmidrule{2-10}
			            & \multirow{2}{*}{MFEA} 
			            & 8154.0 &  449.1 & -- & -- & 635.3 & 717.3 & -- & --\\ 
			            & & 135.34 &  10.59 & -- & -- & 26.98 & 30.54 & -- & --\\ 
			            \cmidrule{2-10}
			            & Wilcoxon test & \tikzcircle[fill=gray]{4pt} & \tikzcircle[fill=gray]{4pt} & -- & -- & \tikzcircle[fill=gray]{4pt} & \tikzcircle[fill=white]{4pt} & -- & -- \\	
			            \midrule
			            
			            \multirow{5}{*}{\rotatebox{90}{\emph{TE\_4\_4}}} & 
			            \multirow{2}{*}{dMFEA-II} 
			            & -- & -- & \textbf{731.4} & \textbf{586.7}  & -- & -- & \textbf{662.9} & \textbf{642.1} \\
			            & & -- & -- & 20.02 & 9.99 & -- & -- & 31.99 & 18.65 \\ 
			            \cmidrule{2-10}
			            & \multirow{2}{*}{MFEA} 
			            & -- & -- & 747.1 & 590.4  & -- & -- & 663.8 & 656.5 \\
			            & & -- & -- & 19.12 & 11.99 & -- & -- & 10.47 & 18.51 \\ 
						\cmidrule{2-10}
			 			& Wilcoxon test & -- & -- & \tikzcircle[fill=white]{4pt} & \tikzcircle[fill=white]{4pt} & -- & -- & \tikzcircle[fill=gray]{4pt} & \tikzcircle[fill=white]{4pt} \\			 			
			            \midrule
			            
			            \multirow{5}{*}{\rotatebox{90}{\emph{TE\_8}}} & 
			            \multirow{2}{*}{dMFEA-II} 
			            & \textbf{8140.8} &  \textbf{451.2} & \textbf{722.7} & \textbf{572.8} & \textbf{614.7} & \textbf{712.1} & \textbf{643.5} & \textbf{642.3}\\ 
			            & & 165.76 & 10.52 & 726.43 & 17.41 & 22.28 & 25.46 & 23.22 & 18.90\\ 
			            \cmidrule{2-10}
			            & \multirow{2}{*}{MFEA} & 
			            8222.5 &  462.0 & 818.1 & 651.5 & 626.9 & 714.6 & 659.2 & 649.5 \\
			            & & 261.57 & 12.76 & 38.11 & 21.67 & 14.78 & 29.72 & 33.90 & 19.98 \\ 
			            \cmidrule{2-10}			            
			            & Wilcoxon test & \tikzcircle[fill=white]{4pt} & \tikzcircle[fill=white]{4pt} & \tikzcircle[fill=white]{4pt} & \tikzcircle[fill=white]{4pt} & \tikzcircle[fill=white]{4pt} & \tikzcircle[fill=gray]{4pt} & \tikzcircle[fill=white]{4pt} & \tikzcircle[fill=gray]{4pt} \\
			            \midrule			            
			            & Optima & 7542 &  426 & 675 & 538 & 554 & 629 & 568 & 598 \\					
						 \\ \bottomrule
		\end{tabular}
	}
	\vspace{2mm}
	\caption{Results obtained by dMFEA-II and MFEA for all the test environments, and statistical significance of the performance gaps as per the Wilcoxon Rank-Sum test.}
	\vspace{-5mm}
	\label{tab:results}
\end{table}

The simulation outputs furnished by the implemented dMFEA-II confirm that this method reaches a better performance than its discrete MFEA counterpart in 22 out of the 24 comparisons that can be established throughout the considered test environments. These findings concur with the conclusions drawn by Bali et al. in \cite{bali2019multifactorial}, namely, that the and learning and adaptation of the parameters driving evolutionary multitasking methods permit to better handle the transfer of negative knowledge, and to leverage even further the existence of synergies among tasks. Going deeper into the results, we observe that dMFEA-II outperforms MFEA in all the eight TSP-VRP instances evolved jointly in \emph{TE\_8}. Furthermore, if we analyze the difference in the results reached in all test environments comprising 4 tasks and in \emph{TE\_8}, dMFEA-II appears to scale better and more resiliently to modifications in the problem instances to solve. Specifically, the results of MFEA degrade significantly when the size of the test environment increases from four to eight simultaneous tasks. Focusing on \texttt{berlin52}, for example, we see that MFEA obtains an average fitness value of 8130.3 in \emph{TE\_4\_1}, 8154.0 in \emph{TE\_4\_2} and a much worse 8222.5 in \emph{TE\_8}. This phenomenon does not occur in dMFEA-II, which maintains its performance in every multitasking environment, even improving it in some cases for \emph{TE\_8}. This is symptomatic of its adaptability, and evinces the superiority of dMFEA-II when compared to the discrete MFEA.

A Wilcoxon Rank-Sum test has been applied over the obtained results to verify the statistical significance of the aforementioned performance gaps. As an example of the analysis conducted in this regard, we comment on the most complex test environment, \emph{TE\_8}. For properly performing this Wilcoxon Rank-Sum test, we have compared the outcomes obtained for all the instances separately, establishing the confidence interval at 90\%. In this way, a white circle ($\tikzcircle[fill=white]{3pt}$) in Table \ref{tab:results} means that dMFEA-II outperforms MFEA with statistical significance. On the contrary, the gray circle ($\tikzcircle[fill=gray]{3pt}$) indicates the non-existence of evidences for ensuring the statistical significance of the performance gap. Thus, the Wilcoxon Rank-Sum test confirms that dMFEA-II significantly outperforms MFEA in 6 of 8 datasets embedded in environment \emph{TE\_8}. Specifically, the obtained average $z$-value is $-2.44$. Considering that the critical $z_c$ value is $-1.64$, and since $-2.44<-1.64$, these results strengthen the significance of the performance differences at $90$\% confidence level. For this reason, we can finally conclude that dMFEA-II is statistically better than MFEA for the multitasking configurations deployed in this experimental study.

\section{Conclusions and Future Work}\label{sec:conc}

This work has presented dMFEA-II, an adaptation of the Multifactorial Evolutionary Algorithm II to permutation-based discrete optimization problems. Specifically, we have elaborated on how the novel ingredients that MFEA-II introduces over its predecessor MFEA have been adapted to deal with solutions encoded as permutations, yielding a new algorithmic proposal that blends together 1) a novel dynamic strategy to update the matrix of evolutionary parameters controlling the exchange of knowledge between tasks; and 2) a new dynamic parent-centric crossover operator suited to deal with permutation-based solutions. For showcasing the application of the proposed dMFEA-II, extensive experiments have been performed using eight different TSP and CVRP instances. We have compared the results attained by dMFEA-II with the ones reached by the discrete variant of MFEA introduced by Yuan et al. in \cite{yuan2016evolutionary}, over five multitasking setups comprising different combinations of the aforementioned problem instances. Results have been conclusive: dMFEA-II outperforms MFEA, with statistical significance, thereby aligning with the claims in \cite{bali2019multifactorial} regarding the intrinsic value of adaptivity in Evolutionary Multitasking. 

Several research directions are planned for the near future departing from the conclusions drawn from this study. First, we will further analyze the scalability of the introduced dMFEA-II using a larger number of TSP and VRP instances. In addition, a critical step is a deeper analysis of the update dynamics of the RMP matrix developed in our dMFEA-II, in order to better understand its behavior along the search process. Finally, we will extrapolate the developed method to other problems arising from other domains with combinatorial optimization at their core.

\section*{Acknowledgments}

Eneko Osaba, Aritz D. Martinez and Javier Del Ser are supported by the Basque Government through the EMAITEK and ELKARTEK funding programs. Javier Del Ser receives support from the Consolidated Research Group MATHMODE (IT1294-19) granted by the Department of Education of the same institution. Andres Iglesias and Akemi Galvez thank the Computer Science National Program of the Spanish Research Agency and European Funds, Project \#TIN2017-89275-R (AEI/FEDER, UE), and the PDE-GIR project of the European Union’s Horizon 2020 programme, Marie Sklodowska-Curie Actions grant agreement \#778035. 

\bibliographystyle{ACM-Reference-Format}
\bibliography{gecco} 

\end{document}